\documentclass[letterpaper]{article} 
\usepackage{aaai2026}  
\usepackage{times}  
\usepackage{helvet}  
\usepackage{courier}  
\usepackage[hyphens]{url}  
\usepackage{graphicx} 
\usepackage{natbib}  
\usepackage{caption} 
\usepackage{algorithm}
\usepackage{algorithmic}

\usepackage{newfloat}
\usepackage{listings}
\DeclareCaptionStyle{ruled}{labelfont=normalfont,labelsep=colon,strut=off} 
\floatstyle{ruled}
\newfloat{listing}{tb}{lst}{}
\floatname{listing}{Listing}

\usepackage{amsmath}
\usepackage{amsfonts}
\usepackage{bm}
\usepackage{multirow}
\usepackage{booktabs}
\title{PatchDecomp: Interpretable Patch-Based Time Series Forecasting}
\author{
    Hiroki Tomioka, 
    Genta Yoshimura\\ 
}
\affiliations{
    Mitsubishi Electric Corporation, Japan\\
    Hiroki.Tomioka@ay.MitsubishiElectric.co.jp
}

\begin{document}

\maketitle

\begin{abstract}
Time series forecasting, which predicts future values from past observations, plays a central role in many domains and has driven the development of highly accurate neural network models.
However, the complexity of these models often limits human understanding of the rationale behind their predictions.
We propose PatchDecomp, a neural network-based time series forecasting method that achieves both high accuracy and interpretability.
PatchDecomp divides input time series into subsequences (patches) and generates predictions by aggregating the contributions of each patch.
This enables clear attribution of each patch, including those from exogenous variables, to the final prediction.
Experiments on multiple benchmark datasets demonstrate that PatchDecomp provides predictive performance comparable to recent forecasting methods.
Furthermore, we show that the model's explanations not only influence predicted values quantitatively but also offer qualitative interpretability through visualization of patch-wise contributions.
\end{abstract}

\begin{links}
    \link{Code}{https://github.com/hiroki-tomioka/PatchDecomp}
\end{links}

\section{Introduction}
Time series data, which record values that change over time, are crucial across various domains, including manufacturing, logistics, and healthcare.
The versatility of time series forecasting (TSF) enables its applications in various domains, such as the prediction electricity consumption, traffic flow, product sales, and inventory levels.
With the advancement of machine learning technologies, numerous forecasting methods have been proposed, ranging from relatively simple linear transformation-based methods to more complex approaches that utilize recurrent neural networks (RNNs), multilayer perceptrons (MLPs), and Transformers~\cite{vaswani2017attention}.
In recent years, there has been a growing trend of incorporating patching techniques that treat input data as subsequences rather than in a point-wise manner, thereby significantly enhancing the performance of TSF.
Additionally, some forecasting frameworks utilize not only the time series history of the target variable but also other exogenous variables (covariates) for predictions.
By incorporating information on external factors that cannot be captured using historical target data alone, these approaches promise further improvements in accuracy, leading to the expansion of these models to real-world problems.

\begin{figure}[t]
    \centering
    \includegraphics[keepaspectratio, width=\linewidth]{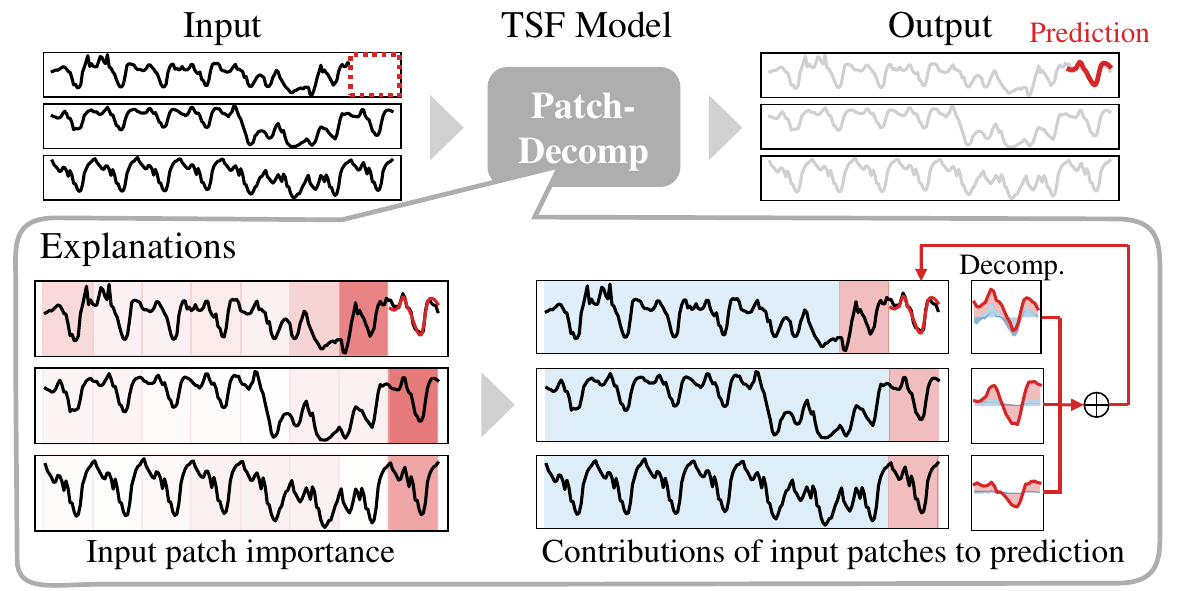}
    \caption{TSF framework and explanations from PatchDecomp. PatchDecomp performs predictions by dividing input time series into patches, providing users with explanations as the input patch importance (the intensity of red in the lower left figure) and the contributions of input patches to the prediction (the area charts in the lower right). For clarity, only the patches with high importance are depicted in red for each variable; others are in blue.}
    \label{fig:key}
\end{figure}

Advancements in studies based on MLPs and Transformers, along with the dramatic increase in computational power, have rapidly improved forecasting accuracy.
However, neural networks are generally regarded as complex black-box models, making it difficult to understand their internal behaviors. This lack of interpretability is undesirable in practical applications.
In real-world settings such as manufacturing systems, users are unlikely to trust predictive models that merely produce outputs without any interpretable rationale.
Deploying such opaque models in domains where transparency and safety are critical entails substantial risks.
Users can assess the validity of these forecasts if the reasons behind the predictions are presented.
This topic is called interpretability and is being actively studied in the field of explainable artificial intelligence (XAI).
For neural network-based TSF methods, enhancing interpretability to mitigate risks and promoting applicability to real-world problems is a significant challenge.
Some existing methods attempt to improve interpretability by visualizing linear weight matrices~\cite{zeng2023transformers}, decomposing predicted values~\cite{oreshkin2019n,olivares2023neural,challu2023nhits}, and presenting variable importance and attention matrix weights over time~\cite{lim2021temporal}.
However, these approaches fail to explain the contributions of each variable's subsequence, including exogenous variables, to the predicted values, rendering their interpretability insufficient for practical use.

In this paper, we propose PatchDecomp, a neural network-based TSF method with interpretability.
PatchDecomp handles the input variables, including exogenous variables, by dividing them into subsequences (patches) and decomposing the predicted values according to the contribution of each input patch (Figure \ref{fig:key}).
By performing contribution decomposition based on the entire processing of the model from input to output, this model is capable of rigorously calculating the correspondence between inputs and outputs, rather than being limited to partial interpretability through the visualization of the attention map.
This model is designed to improve interpretability, and achieving state-of-the-art performance is not the focus of this study.
However, as a result of extensive experiments, we confirmed the competitive predictive accuracy of PatchDecomp.
The main contributions are summarized as follows:
\begin{itemize}
    \item We propose an interpretable patch-based TSF model called PatchDecomp. This model can deal with exogenous variables in addition to the target and explain the contribution of each variable's patch to the prediction.
    \item We conduct TSF tasks on multiple datasets and demonstrate that the proposed method is comparable to recent forecasting methods in terms of its predictive accuracy.
    \item By conducting both qualitative and quantitative analyses, we demonstrate that the proposed method achieves superior interpretability in TSF.
\end{itemize}

\section{Related Work}

\subsection{Time Series Forecasting}
Although various TSF methods have been studied for a long time, deep learning models have become mainstream.
Recently, models based on MLPs and Transformers have largely replaced RNN-based models, such as DeepAR~\cite{salinas2020deepar}, for TSF trends.
N-BEATS~\cite{oreshkin2019n} is an MLP-based method that stacks predicted trend and seasonal components.
NBEATSx~\cite{olivares2023neural} was developed subsequently to handle exogenous variables, along with NHITS~\cite{challu2023nhits}, which is hierarchical and capable of processing multiple frequencies.
TSMixer~\cite{chen2023tsmixer}, which applies MLP-Mixer, and TiDE~\cite{das2023long}, which features an encoder--decoder structure using MLPs, can handle exogenous variables.
Moreover, Transformer fully utilizes attention mechanisms, and the development of models incorporating these architectures, such as Informer~\cite{zhou2021informer}, Autoformer~\cite{wu2021autoformer}, and FEDformer~\cite{zhou2022fedformer}, is one of the most popular directions.
Temporal Fusion Transformer (TFT)~\cite{lim2021temporal} consists of an LSTM encoder and a multi-head attention decoder and utilizes exogenous variables to predict future time series.
While conventional Transformer-based methods tokenize data by time steps and apply attention, iTransformer~\cite{liu2023itransformer} represents a paradigm shift by tokenizing data by variables instead of time.

Furthermore, a noteworthy idea is the patching technique introduced by PatchTST~\cite{nie2022time}, which enables the model to capture local time series information that cannot be grasped point-wise by splitting the input time series into patches before feeding them into the Transformer encoder.
The patching technique has had a significant influence on subsequent research and has become essential for enhancing forecasting models.
TimeXer~\cite{wang2024timexer} also adopts this technique and comprises patch-wise self-attention and cross-attention with exogenous variables.

The proposed method herein adopts the attention mechanism and also employs a patching technique to forecast future values based on the patches of the input variables.

\subsection{Interpretability of TSF Models}
In the modern era, where AI technologies are spreading throughout society, understanding the behavior of AI models is critically important and is being actively studied across a wide range of areas within machine learning.
Several researchers are working on time series tasks~\cite{zhao2023interpretation,ozyegen2022evaluation,arsenault2025survey}; however, much of this work remains limited to time series classification problems.

Studies on the interpretability of time series tasks can be broadly divided into two categories: post-hoc interpretation methods and inherently interpretable models~\cite{zhao2023interpretation}.

Post-hoc interpretation methods extract interpretable information by applying it to pretrained models.
LIME~\cite{ribeiro2016should} and SHAP~\cite{lundberg2017unified} fall into this category.
Although these methods can be applied independently to the predictive model, they have the drawbacks of relatively high computational costs and difficulty providing precise explanations of the model behavior because they perform an approximation.

On the other hand, inherently interpretable models have build-in mechanisms to present the rationale behind their predictions.
While it is necessary to incorporate explanatory mechanisms into the model, explanations that are more understandable to users (compared with post-hoc methods) can be provided with careful design.
For example, simple models, such as DLinear and NLinear~\cite{zeng2023transformers} possess interpretability in their linear weight matrices.
By visualizing these weights, users can understand the variables that strongly contribute to predictions.
In addition, N-BEATS, NBEATSx, and NHITS, can decompose the predicted time series into trend and seasonal components, allowing us to understand which variables or components contribute to the forecasted values.
TFT enhances interpretability by calculating the attention weights for each time step and the feature importance for each variable.
Transformer-based methods often regard attention weights as a foundation for interpretability, as visualizing these weights provides insights into which inputs are more focused by the model~\cite{schockaert2020attention,gangopadhyay2021spatiotemporal,liu2023itransformer}.
However, in the context of natural language processing, while attention weights provide interpretation to some extent, they do not possess sufficient capability to fully explain model behavior~\cite{serrano2019attention,jain2019attention}, and the effectiveness of attention weights in interpreting TSF models remains a controversial topic.

\begin{figure*}[t]
    \centering
    \includegraphics[keepaspectratio, width=0.95\linewidth]{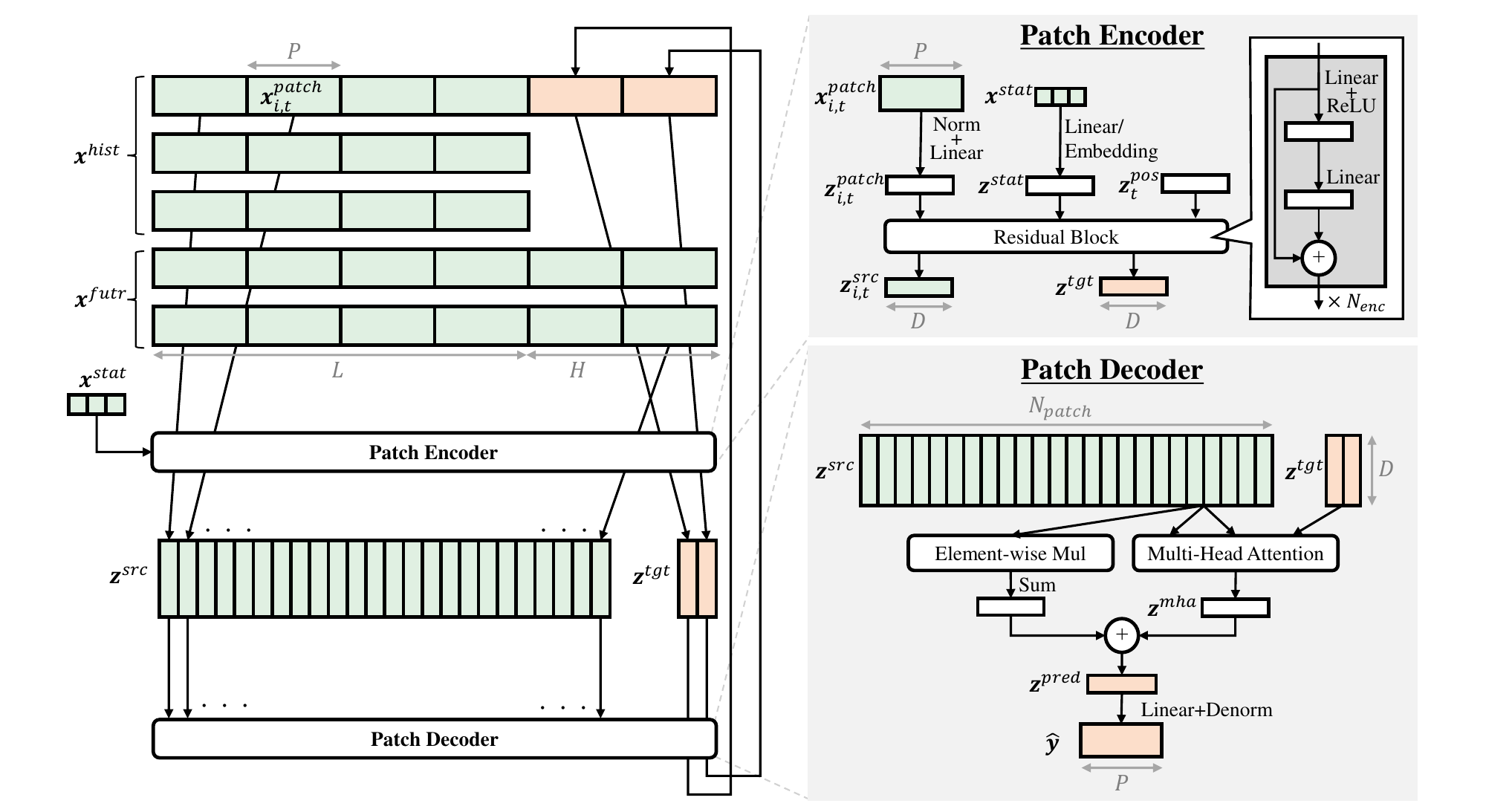}
    \caption{Architecture of PatchDecomp}
    \label{fig:architecture}
\end{figure*}

In all the aforementioned models, the extent to which each segment of the variables, including exogenous variables, contributes to the predicted values remains unclear.
In this paper, we propose an inherently interpretable model that decomposes the contributions of input patches to predicted values.
Our approach does not simply visualize attention weights; rather, it establishes an apparent correspondence between input and output data by considering the entire processing of the architecture that includes the attention mechanism.

\section{Methodology}

In this study, we address the problem of predicting future time series values over $H$ time steps, based on the observed time series history over $L$ time steps.
The current time series history of the target variable $\bm{y}$ is denoted as $\bm{y}_{:L}=\{y_1,\cdots,y_L\}$, whereas the future time series we aim to predict is represented as $\bm{y}_{-H:}=\{y_{L+1},\cdots,y_{L+H}\}$.

We denote $D_\text{hist}$ exogenous variables that can be observed from the current time series history, along with the target variable $\bm{y}_{:L}$, as $\bm{x}^{\text{hist}} \in \mathbb{R}^{(D_\text{hist} + 1) \times L}$.
We denote $D_\text{futr}$ exogenous variables, such as weather forecasts and calendar information, for which the time series is known in advance up to the future time steps, as $\bm{x}^{\text{futr}} \in \mathbb{R}^{D_\text{futr} \times (L+H)}$.
In addition, we describe static variables that do not change over time, such as product IDs and categories, as $\bm{x}^\text{stat} \in \mathbb{R}^{D_\text{stat}}$.

The objective of TSF is to find a model $\mathcal{F}$ that can accurately predict the future time series $\bm{y}_{-H:}$ using exogenous variables as inputs, as follows:
\begin{equation}
    \hat{\bm{y}}_{-H:}=\mathcal{F}(\bm{x}^\text{hist},\bm{x}^\text{futr},\bm{x}^\text{stat}) \in \mathbb{R}^{H},
\end{equation}
where $\hat{\bm{y}}_{-H:}$ represents the predicted future values over $H$ time steps.

As shown in Figure \ref{fig:architecture}, PatchDecomp consists of two components: a patch encoder, which divides the time series into patches for each variable and encodes them into latent vectors, and a patch decoder, which associates the latent vectors of the input and output patches and decodes them into predicted values.

\subsection{Patch Encoder}
First, the time series is standardized using reversible instance normalization (RevIN)~\cite{kim2021reversible} for each variable and then divided into patches of length $P$, starting from the current time.
If the past length $L$ or future length $H$ is not divisible by the patch length $P$, zero padding is applied at the beginning or end.
Consequently, the numbers of past and future patches are given by $N_\text{hist}=\lceil\frac{L}{P}\rceil$ and $N_\text{futr}=\lceil\frac{H}{P}\rceil$, respectively.

Next, for the $i$-th variable's $t$-th patch $\bm{x}^\text{patch}_{i,t} \in \mathbb{R}^P$, a linear transformation is applied to obtain $\bm{z}^\text{patch}_{i,t} \in \mathbb{R}^D$.
Temporal information is encoded via positional encoding through linear transformation or embedding, resulting in $\bm{z}^\text{pos}_t \in \mathbb{R}^{(N_\text{hist} + N_\text{futr}) \times D}$.
Additionally, to embed the information from the static variables $\bm{x}^\text{stat} \in \mathbb{R}^{D_\text{stat}}$, a linear transformation or embedding is applied to obtain $\bm{z}^\text{stat} \in \mathbb{R}^D$, which is also added.

Subsequently, by applying a residual block that combines the MLP and skip connections $N_\text{enc}$ times, each patch is encoded into a $D$-dimensional latent vector while considering the temporal positional information and static variables:
\begin{equation}
    \bm{z}^\text{src}_{i,t}=\text{Encode}(\bm{z}^\text{patch}_{i,t}+\bm{z}^\text{pos}_t+\bm{z}^\text{stat}) \in \mathbb{R}^D.
\end{equation}
The resulting $\bm{z}^\text{src} \in \mathbb{R}^{N_\text{patch} \times D}$, which is formed by stacking the vectors $\bm{z}^\text{src}_{i,t}$ for all patches, corresponds to the latent representation for the input, where $N_\text{patch}=(1+D_\text{hist}+D_\text{futr})N_\text{hist}+D_\text{futr}N_\text{futr}$ denotes the total number of patches.

Furthermore, the latent representation corresponding to the output can be obtained by encoding similarly, excluding $\bm{z}^\text{patch}_{i,t}$:
\begin{equation}
    \bm{z}^\text{tgt}_t=\text{Encode}(\bm{z}^\text{pos}_t+\bm{z}^\text{stat}) \in \mathbb{R}^D.
\end{equation}
$\bm{z}^\text{tgt} \in \mathbb{R}^{N_\text{futr} \times D}$ is also acquired by stacking $\bm{z}^\text{tgt}_t$.

\subsection{Patch Decoder}
By using the representations of the output patches $\bm{z}^\text{tgt}$ as the query and the representations of the input patches $\bm{z}^\text{src}$ as the key and value in a multi-head attention mechanism with $N_\text{head}$ heads, it is possible to relate the contributions of the input patches to the output patches in a decomposable manner:
\begin{equation}
    \bm{z}^\text{mha} = \text{MultiHeadAttention}(\bm{z}^\text{tgt}, \bm{z}^\text{src}, \bm{z}^\text{src}) \in \mathbb{R}^{N_\text{futr} \times D}.
\end{equation}

We obtain the latent vector for the patches corresponding to the prediction output by adding the sum of the element-wise product of the latent representation of the input patches $\bm{z}^\text{src}$ and the bias vector $\bm{w}^\text{bias} \in \mathbb{R}^{N_\text{patch} \times D}$:
\begin{equation}
    \bm{z}^\text{pred} = \bm{z}^\text{mha} + \text{Sum}(\bm{z}^\text{src} * \bm{w}^\text{bias}) \in \mathbb{R}^{N_\text{futr} \times D}.
\end{equation}

Finally, we obtain the predicted values $\hat{\bm{y}}_{-H:}$ by applying a linear transformation to $\bm{z}^\text{pred}$ to convert it into the patch dimension and then reverting it to the original scale using the mean and standard deviation calculated by RevIN.

\subsection{Decomposition of the prediction}
In the patch encoder, the $i$-th variable's $t$-th patch $\bm{x}^\text{patch}_{i,t}$ is independently mapped to a representation vector $\bm{z}^\text{patch}_{i,t}$.
On the other hand, in the patch decoder, we utilize multi-head attention only once to associate the input and output patches and subsequently apply linear transformations.
When the dimension of each head's value is $d_v$, multi-head attention typically computes attention via matrix multiplication between an attention weight of dimension $N_\text{futr} \times N_\text{patch}$ and a value of dimension $N_\text{patch} \times d_v$.
However, we first compute the element-wise product of tensors of dimension $N_\text{futr} \times N_\text{patch} \times 1$ and $1 \times N_\text{patch} \times d_v$, then sum along the $N_\text{patch}$ dimension.
This method calculates the same attention while providing the contributions along the $N_\text{patch}$ dimension.
Additionally, the bias vectors are also computed for each patch.
Thus, this architecture can decompose the output according to the contributions of each input patch through the series of processes in the encoder and decoder.
This provides the model's predictive rationale more directly than the visualization of attention maps.


\section{Experiments}

\subsection{Experimental Settings}

\subsubsection{Datasets.}
We conducted long-term TSF (LTSF) using seven datasets frequently used as benchmarks: ETTh1, ETTh2, ETTm1, ETTm2, Weather, Electricity (ECL), and Traffic.
These datasets did not contain exogenous variables, and in the experimental setup, each dataset was considered a univariate time series.
By contrast, to evaluate the performance in TSF tasks that include exogenous variables, we used the electricity price forecasting (EPF) datasets~\cite{lago2021forecasting}, which contain real data from five electricity markets (NP, PJM, BE, FR, and DE).
This dataset included two different exogenous variables for each electricity market, such as system load and wind generation.
In addition, we used the month, day of the week, and hour as exogenous variables.

\subsubsection{Baselines.}
As the baseline of LTSF without exogenous variables, we employed the following models: PatchTST~\cite{nie2022time}, NBEATSx~\cite{olivares2023neural}, NHITS~\cite{challu2023nhits}, TFT~\cite{lim2021temporal}, DLinear~\cite{zeng2023transformers}, TSMixer~\cite{chen2023tsmixer}, Autoformer~\cite{wu2021autoformer}, iTransformer~\cite{liu2023itransformer}, and TiDE~\cite{das2023long}.
For the EPF task, we selected models capable of handling exogenous variables, specifically NBEATSx, NHITS, TFT, TSMixer, and TiDE.

\subsubsection{Experimental Details.}
The proposed and existing methods were implemented and evaluated using NeuralForecast~\cite{olivares2022library_neuralforecast}.
In all the experiments, the learning rate was set to $10^{-3}$, and the loss function was the mean absolute error (MAE).
The data were split into training, validation, and test sets in chronological order to ensure no overlap between the sets.
The training was terminated when the prediction accuracy on the validation set no longer improved (early stopping).
The hyperparameters that yielded the best prediction accuracy in the validation set were explored using Optuna~\cite{akiba2019optuna}.
Although the patch length $P$ can be set in two ways: fixed and variable length, variable length patches do not necessarily align with the user's intuition.
For example, calculating contributions for patches of length 17 hours for hourly data could potentially hinder the user's understanding.
Considering the affinity with interpretability, this study adopted fixed-length patches.
For the LTSF task, the optimal value was determined through hyperparameter tuning from $P\in\{12, 24, 48\}$, whereas $P$ was specifically set to 24 to align with daily units for the EPF task.
All the experiments under each condition were conducted in five trials using different seed values on an NVIDIA Quadro RTX8000 GPU.

\begin{table*}[ht]
    \centering
    \small
    \begin{tabular}{c|c|cc|cc|cc|cc|cc|cc|cc|cc|cc|cc}
        \toprule
        \multicolumn{2}{c|}{} & \multicolumn{2}{c|}{PatchDecomp} & \multicolumn{2}{c|}{PatchTST} & \multicolumn{2}{c|}{NBEATSx} & \multicolumn{2}{c|}{NHITS} & \multicolumn{2}{c|}{TFT} & \multicolumn{2}{c|}{DLinear} & \multicolumn{2}{c|}{TSMixer} & \multicolumn{2}{c|}{Autoformer} & \multicolumn{2}{c|}{iTransformer} & \multicolumn{2}{c}{TiDE} \\
        \midrule
        \multicolumn{2}{c|}{} & MSE & MAE & MSE & MAE & MSE & MAE & MSE & MAE & MSE & MAE & MSE & MAE & MSE & MAE & MSE & MAE & MSE & MAE & MSE & MAE \\
        \midrule
        \multirow{4}{*}{\rotatebox[origin=c]{90}{ETTh1}} & 96 & \underline{0.362} & \underline{0.412} & 0.405 & 0.436 & 0.658 & 0.550 & 0.703 & 0.555 & 0.610 & 0.519 & 0.380 & 0.424 & 0.435 & 0.459 & 0.513 & 0.497 & 0.432 & 0.459 & \textbf{0.356} & \textbf{0.408} \\
        & 192 & \textbf{0.383} & \textbf{0.427} & 0.415 & 0.443 & 0.730 & 0.582 & 0.758 & 0.583 & 0.648 & 0.544 & \underline{0.404} & \underline{0.442} & 0.465 & 0.477 & 0.507 & 0.489 & 0.491 & 0.495 & 0.418 & 0.450 \\
        & 336 & \textbf{0.383} & \textbf{0.431} & 0.439 & \underline{0.461} & 0.809 & 0.619 & 0.792 & 0.604 & 0.623 & 0.545 & \underline{0.442} & 0.467 & 0.528 & 0.520 & 0.485 & 0.492 & 0.517 & 0.518 & 0.462 & 0.482 \\
        & 720 & \textbf{0.406} & \textbf{0.453} & 0.457 & 0.483 & 0.871 & 0.655 & 0.843 & 0.631 & 0.624 & 0.562 & \underline{0.440} & \underline{0.476} & 0.538 & 0.538 & 0.543 & 0.538 & 0.548 & 0.547 & 0.457 & 0.488 \\
        \midrule
        \multirow{4}{*}{\rotatebox[origin=c]{90}{ETTh2}} & 96 & \underline{0.264} & \textbf{0.339} & \textbf{0.263} & \underline{0.340} & 0.319 & 0.373 & 0.330 & 0.375 & 0.286 & 0.363 & 0.268 & 0.346 & 0.353 & 0.413 & 0.327 & 0.396 & 0.373 & 0.427 & 0.275 & 0.351 \\
        & 192 & \textbf{0.316} & \textbf{0.374} & 0.321 & \underline{0.379} & 0.395 & 0.421 & 0.409 & 0.421 & 0.337 & 0.393 & \underline{0.318} & 0.380 & 0.412 & 0.450 & 0.343 & 0.406 & 0.408 & 0.449 & 0.325 & 0.384 \\
        & 336 & \underline{0.341} & \textbf{0.395} & 0.354 & 0.406 & 0.451 & 0.458 & 0.461 & 0.457 & 0.360 & 0.410 & 0.343 & \underline{0.401} & 0.432 & 0.465 & \textbf{0.336} & 0.404 & 0.443 & 0.474 & 0.350 & 0.405 \\
        & 720 & \textbf{0.366} & \textbf{0.422} & 0.372 & 0.424 & 0.485 & 0.485 & 0.477 & 0.474 & 0.379 & 0.432 & \underline{0.367} & \underline{0.423} & 0.484 & 0.501 & 0.381 & 0.439 & 0.483 & 0.502 & 0.374 & 0.427 \\
        \midrule
        \multirow{4}{*}{\rotatebox[origin=c]{90}{ETTm1}} & 96 & 0.567 & 0.501 & 0.322 & \underline{0.374} & 0.528 & 0.479 & 0.560 & 0.485 & 0.610 & 0.504 & \textbf{0.310} & \textbf{0.368} & 0.439 & 0.441 & 0.582 & 0.511 & 0.443 & 0.452 & \underline{0.321} & 0.378 \\
        & 192 & 0.569 & 0.502 & 0.356 & 0.394 & 0.602 & 0.520 & 0.569 & 0.502 & 0.639 & 0.515 & \textbf{0.333} & \textbf{0.383} & 0.448 & 0.448 & 0.594 & 0.519 & 0.470 & 0.465 & \underline{0.345} & \underline{0.393} \\
        & 336 & 0.567 & 0.502 & 0.384 & \underline{0.409} & 0.669 & 0.549 & 0.655 & 0.536 & 0.619 & 0.519 & \textbf{0.353} & \textbf{0.395} & 0.467 & 0.459 & 0.598 & 0.522 & 0.487 & 0.476 & \underline{0.362} & \underline{0.403} \\
        & 720 & 0.595 & 0.523 & 0.441 & 0.445 & 0.767 & 0.599 & 0.730 & 0.571 & 0.629 & 0.530 & \textbf{0.397} & \textbf{0.420} & 0.506 & 0.480 & 0.619 & 0.536 & 0.534 & 0.502 & \underline{0.410} & \underline{0.430} \\
        \midrule
        \multirow{4}{*}{\rotatebox[origin=c]{90}{ETTm2}} & 96 & 0.190 & 0.290 & \textbf{0.166} & \textbf{0.259} & 0.183 & 0.279 & 0.186 & 0.278 & 0.203 & 0.298 & \underline{0.168} & \underline{0.265} & 0.340 & 0.379 & 0.256 & 0.339 & 0.340 & 0.380 & 0.173 & 0.269 \\
        & 192 & 0.230 & 0.315 & \textbf{0.215} & \textbf{0.294} & 0.245 & 0.322 & 0.249 & 0.320 & 0.245 & 0.327 & \textbf{0.215} & \underline{0.297} & 0.400 & 0.411 & 0.279 & 0.352 & 0.388 & 0.406 & 0.220 & 0.302 \\
        & 336 & 0.275 & 0.345 & \textbf{0.261} & \textbf{0.328} & 0.302 & 0.362 & 0.307 & 0.358 & 0.284 & 0.351 & \textbf{0.261} & \underline{0.329} & 0.439 & 0.433 & 0.310 & 0.371 & 0.434 & 0.432 & 0.265 & 0.331 \\
        & 720 & 0.349 & 0.390 & 0.337 & \underline{0.377} & 0.392 & 0.419 & 0.396 & 0.412 & 0.352 & 0.395 & \textbf{0.334} & \textbf{0.376} & 0.494 & 0.471 & 0.369 & 0.408 & 0.481 & 0.465 & \underline{0.336} & 0.378 \\
        \midrule
        \multirow{4}{*}{\rotatebox[origin=c]{90}{Weather}} & 96 & \textbf{0.171} & 0.212 & 0.176 & \textbf{0.198} & 0.196 & \underline{0.200} & 0.199 & 0.201 & \underline{0.174} & 0.214 & 0.179 & 0.210 & 0.340 & 0.325 & 0.254 & 0.286 & 0.369 & 0.343 & 0.184 & 0.218 \\
        & 192 & \textbf{0.215} & 0.248 & 0.219 & \textbf{0.236} & 0.248 & 0.243 & 0.251 & \underline{0.242} & \underline{0.216} & 0.246 & 0.223 & 0.246 & 0.392 & 0.351 & 0.277 & 0.298 & 0.424 & 0.371 & 0.228 & 0.251 \\
        & 336 & \textbf{0.264} & 0.281 & 0.269 & \textbf{0.273} & 0.299 & 0.284 & 0.301 & 0.283 & \underline{0.267} & 0.281 & \underline{0.267} & \underline{0.279} & 0.420 & 0.370 & 0.300 & 0.311 & 0.451 & 0.389 & 0.271 & 0.283 \\
        & 720 & \textbf{0.316} & \textbf{0.313} & 0.322 & \textbf{0.313} & 0.360 & 0.330 & 0.362 & 0.326 & \underline{0.318} & 0.314 & 0.322 & 0.316 & 0.413 & 0.373 & 0.333 & 0.329 & 0.432 & 0.384 & 0.325 & 0.320 \\
        \midrule
        \multirow{4}{*}{\rotatebox[origin=c]{90}{ECL}} & 96 & 0.148 & \underline{0.237} & 0.149 & 0.250 & 0.158 & 0.257 & 0.155 & 0.253 & 0.281 & 0.346 & \textbf{0.140} & \textbf{0.236} & 0.218 & 0.309 & 0.266 & 0.333 & 0.237 & 0.332 & \underline{0.141} & \underline{0.237} \\
        & 192 & 0.168 & 0.254 & 0.159 & 0.262 & 0.175 & 0.272 & 0.170 & 0.268 & 0.398 & 0.421 & \textbf{0.149} & \textbf{0.245} & 0.237 & 0.324 & 0.310 & 0.364 & 0.259 & 0.349 & \textbf{0.149} & \underline{0.246} \\
        & 336 & 0.181 & 0.269 & 0.171 & 0.277 & 0.188 & 0.289 & 0.184 & 0.284 & 0.831 & 0.674 & \underline{0.160} & \underline{0.261} & 0.253 & 0.338 & 0.285 & 0.353 & 0.272 & 0.361 & \textbf{0.158} & \textbf{0.258} \\
        & 720 & 0.242 & 0.318 & 0.221 & 0.311 & 0.231 & 0.326 & 0.222 & 0.319 & 0.910 & 0.710 & \textbf{0.198} & \textbf{0.296} & 0.321 & 0.392 & 0.476 & 0.479 & 0.332 & 0.405 & \underline{0.200} & \underline{0.298} \\
        \midrule
        \multirow{4}{*}{\rotatebox[origin=c]{90}{Traffic}} & 96 & 0.418 & \textbf{0.289} & 0.399 & 0.312 & \textbf{0.381} & \underline{0.300} & \underline{0.383} & 0.301 & 0.528 & 0.346 & 0.417 & 0.305 & 0.453 & 0.326 & 0.611 & 0.428 & 0.439 & 0.352 & 0.415 & 0.302 \\
        & 192 & 0.431 & \textbf{0.294} & 0.426 & 0.319 & \underline{0.409} & 0.315 & \textbf{0.408} & 0.315 & 0.610 & 0.392 & 0.433 & 0.311 & 0.470 & 0.332 & 0.631 & 0.442 & 0.457 & 0.359 & 0.428 & \underline{0.308} \\
        & 336 & 0.433 & \textbf{0.297} & \underline{0.425} & 0.322 & \underline{0.425} & 0.323 & \textbf{0.423} & 0.322 & 0.690 & 0.429 & 0.440 & 0.317 & 0.473 & 0.333 & 0.572 & 0.402 & 0.467 & 0.364 & 0.438 & \underline{0.315} \\
        & 720 & 0.480 & \textbf{0.329} & \underline{0.472} & 0.353 & 0.477 & 0.356 & \textbf{0.463} & 0.348 & 1.063 & 0.646 & 0.486 & 0.348 & 0.584 & 0.376 & 0.767 & 0.510 & 0.564 & 0.402 & 0.482 & \underline{0.344} \\
        \bottomrule
    \end{tabular}
    \caption{LTSF results. The best results are in bold, and the second best are underlined.}
    \label{tab:ltsf}
\end{table*}

\subsection{Accuracy}
In the LTSF task, the past time series length was set to $L=512$, and the future length was configured to four options: $H \in \{96, 192, 336, 720\}$.
The quantitative prediction accuracies of the proposed and baseline methods are summarized in Table \ref{tab:ltsf}.

The performance varied across the seven datasets; however, PatchDecomp recorded high prediction accuracies on several datasets.
Notably, impressive results were obtained for ETTh1 and ETTh2.
These results can be attributed to the fact that ETTh1 and ETTh2 exhibit significantly pronounced periodicity below the patch length compared with the other datasets, which aligns well with the characteristics of PatchDecomp, which utilizes patch segmentation for forecasting.

Table \ref{tab:epf} presents the results for EPF tasks.
In this experiment, the past length was set to $L=168$ and the future length was set to $H=24$.
Similar to the LTSF task, the results for the EPF task exhibited variability depending on the data source (market).
However, PatchDecomp and TFT tended to achieve higher accuracies.

\begin{table*}[htbp]
    \centering
    \small
    \begin{tabular}{c|cc|cc|cc|cc|cc|cc}
        \toprule
         & \multicolumn{2}{c|}{PatchDecomp} & \multicolumn{2}{c|}{NBEATSx} & \multicolumn{2}{c|}{NHITS} & \multicolumn{2}{c|}{TFT} & \multicolumn{2}{c|}{TSMixer} & \multicolumn{2}{c}{TiDE} \\
        \midrule
         & MSE & MAE & MSE & MAE & MSE & MAE & MSE & MAE & MSE & MAE & MSE & MAE \\
        \midrule
        NP & 18.32 & 2.26 & 18.31 & 2.27 & \underline{17.86} & \underline{2.22} & \textbf{16.72} & \textbf{2.11} & 46.33 & 4.09 & 27.20 & 2.95 \\
        PJM & \textbf{28.81} & \textbf{3.23} & \underline{30.88} & 3.42 & 31.88 & \underline{3.39} & 32.42 & 3.40 & 78.66 & 5.50 & 44.15 & 4.19 \\
        BE & \textbf{182.93} & \textbf{5.36} & \underline{184.70} & 5.54 & 186.74 & 5.60 & 187.65 & \underline{5.37} & 321.08 & 8.29 & 217.41 & 6.57 \\
        FR & \underline{218.42} & \underline{4.70} & 219.09 & 4.72 & 218.75 & 4.73 & \textbf{215.81} & \textbf{4.59} & 341.97 & 7.60 & 242.97 & 5.80 \\
        DE & 65.18 & 4.96 & 61.83 & 4.80 & \underline{59.15} & \underline{4.68} & \textbf{58.99} & \textbf{4.58} & 211.49 & 9.15 & 139.03 & 7.63 \\
        \bottomrule
    \end{tabular}
    \caption{EPF results. The best results are in bold, and the second best are underlined.}
    \label{tab:epf}
\end{table*}

\begin{figure}[t]
    \centering
    \includegraphics[keepaspectratio, width=\linewidth]{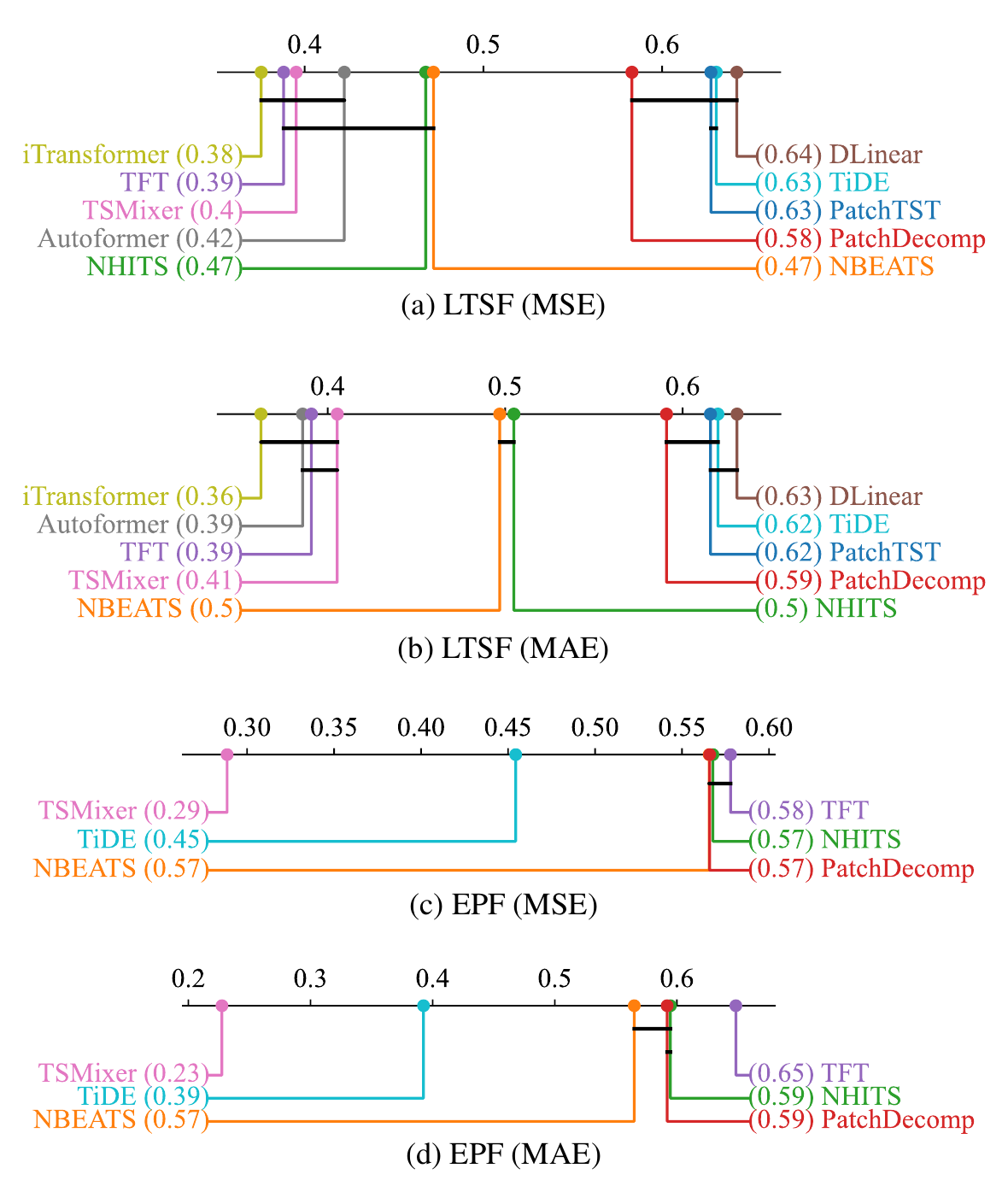}
    \caption{Critical difference diagrams. The horizontal axis corresponds to the performance rankings of each method, with methods positioned further to the right on the graph indicating a relatively higher performance. The black crossbars connect methods that do not exhibit statistically significant differences.}
    \label{fig:CD}
\end{figure}

To statistically verify the performance differences between methods for the LTSF and EPF tasks, critical difference diagrams~\cite{demvsar2006statistical} were created (Figure \ref{fig:CD}).
The Conover's test~\cite{conover1979multiple}, a type of nonparametric test, was utilized for post-hoc analysis.
In the evaluation of the LTSF based on the mean squared error (MSE) (Figure \ref{fig:CD}(a)), the proposed method statistically belonged to the highest-ranking group, and it was ranked in the second highest group when evaluated based on MAE (Figure \ref{fig:CD}(b)).
Similar results were observed for the EPF task as well (Figure \ref{fig:CD}(c) and (d)).
Thus, PatchDecomp does not fall short of the baseline methods in terms of prediction accuracy.

\subsection{Interpretability}
Focusing on the EPF task with exogenous variables, we demonstrate the interpretability of PatchDecomp from both qualitative and quantitative perspectives.
As necessary, we clarify the characteristics of the proposed method by comparing it with TFT, which provides interpretability in terms of feature importance and attention weights.
For TFT, the importance of a specific variable at a given time can be calculated on a point-wise basis by combining these two types of characteristics.

\subsubsection{Qualitative Evaluation.}
\begin{figure}[htbp]
    \centering
    \includegraphics[keepaspectratio, width=\linewidth]{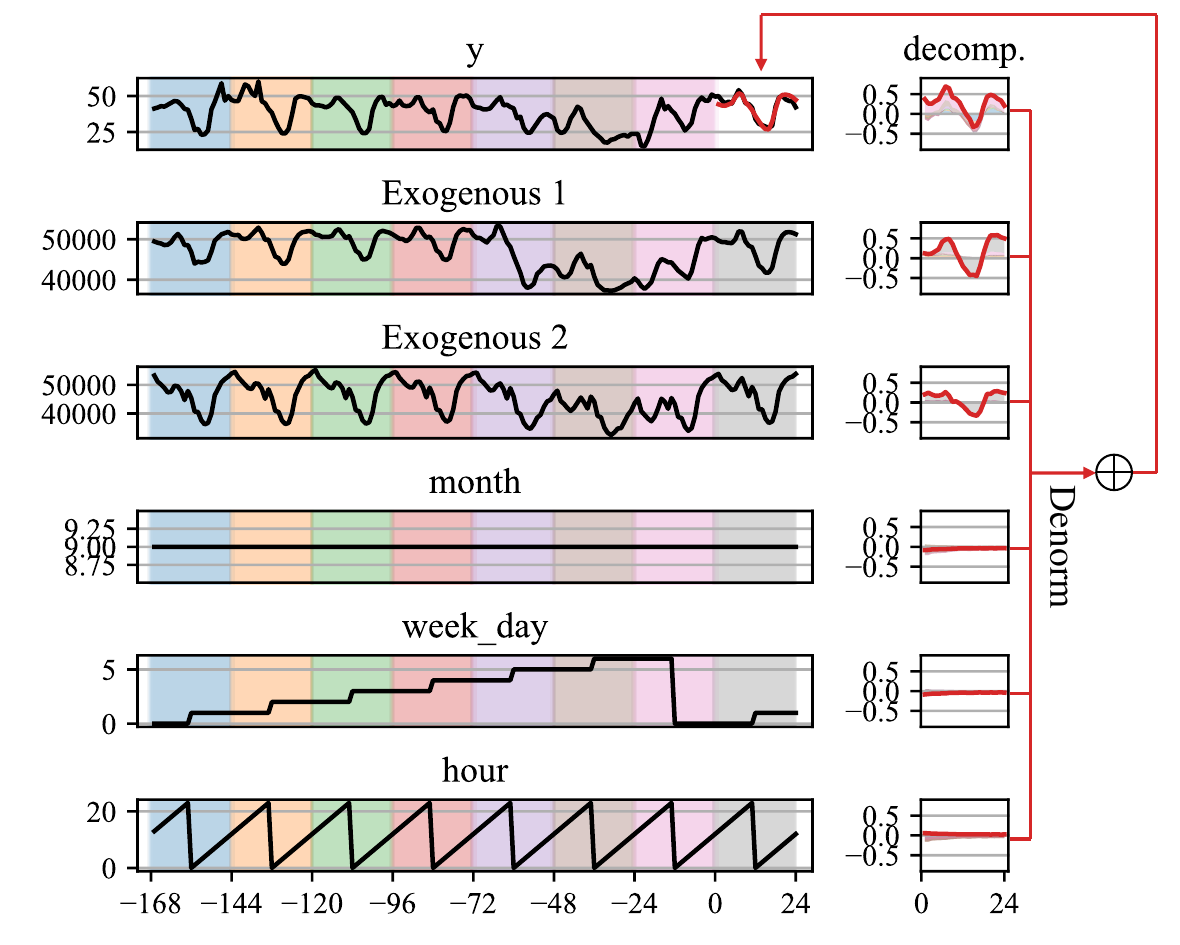}
    \caption{Decomposition of the contributions of the input patches. Each row represents a variable, with the colors corresponding to the input patches. The right column displays the area charts of the prediction contributions for each variable, where the final prediction is obtained by summing the contributions of all the variables.}
    \label{fig:decomposition}
\end{figure}

\begin{figure}[htbp]
    \centering
    \includegraphics[keepaspectratio, width=\linewidth]{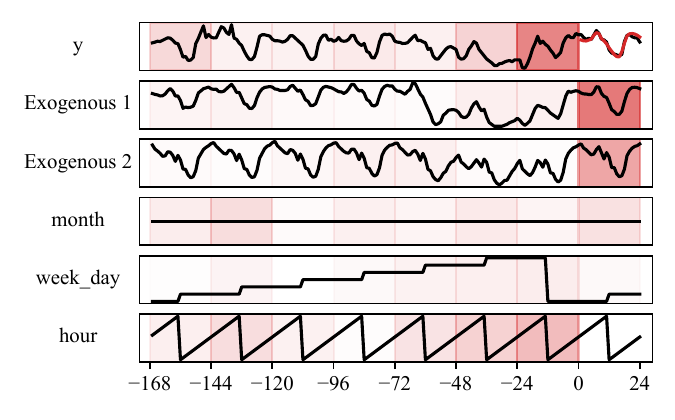}
    \caption{Local explanation. It represents the prediction contributions for each input patch at a specific time by the intensity of the color.}
    \label{fig:local_explanation}
\end{figure}

\begin{figure}[htbp]
    \centering
    \includegraphics[keepaspectratio, width=\linewidth]{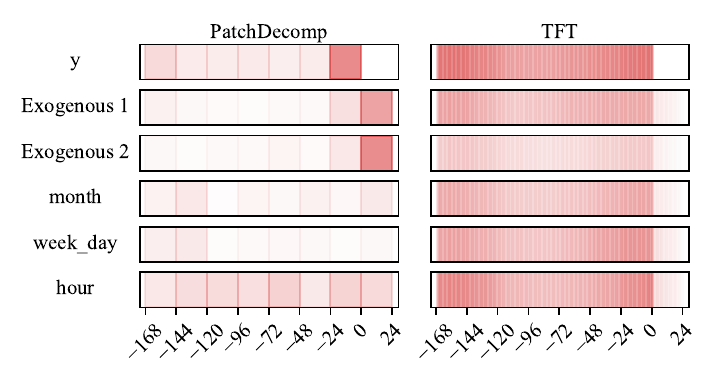}
    \caption{Global explanations of PatchDecomp and TFT. The darker the color, the higher the importance of that part across the entire test dataset.}
    \label{fig:global_explanations}
\end{figure}

The key aspect of the interpretability of PatchDecomp lies in its ability to decompose predicted values based on the contributions of each input patch.
Figure \ref{fig:decomposition} shows an example of the decomposition results for the predicted values in the BE market.
The x-axis represents time, and 0 indicates the current time point.
The first row (y) shows the time series of the target electricity price, where the black and red lines represent the actual and predicted values, respectively.
The subsequent rows show the exogenous variables used for the prediction, listed from top to bottom: system load (exogenous 1), generation (exogenous 2), month, day of the week (week\_day), and hour.
In the right column (decomp.), the contributions of each variable to the predicted values are indicated by red lines, and the sum of the red lines for each variable aligns with the final predicted value.
Moreover, for each variable, the contributions of the individual patches are displayed as an area chart, with the background color of the patches corresponding to the respective colors in the chart.
This enables a precise understanding of the extent to which data contained in the input patches influenced the actual predicted values.
Consequently, it is easier to identify the dominant input data affecting the predictions and pinpoint the data responsible for any unnatural predictions, which is expected to significantly contribute to practical applications.
For instance, in this case, we can interpret that the patch immediately before the prediction (pink) and the future  system load and generation (exogenous 1 and 2, gray) strongly contribute to the prediction.
Patch-level explanations, rather than point-wise explanations, can become a user-friendly approach in multi-horizon forecasting problems, as they allow hourly data to be interpreted at granularities such as ``data from yesterday" or ``data from the same day of the previous week."
Additionally, Figure \ref{fig:local_explanation} shows a local explanation for a specific time-point prediction, where the contribution of each patch is represented by the intensity of the color.
This intensity was calculated from the sum of the absolute values (areas) of the contributions of each patch, as shown in Figure \ref{fig:decomposition}.
Figure \ref{fig:global_explanations} shows global explanations that illustrate the prediction contributions of each patch across the entire test dataset for both the proposed method and TFT.
Global explanations elucidate the behavior of the model for the target dataset.
When comparing the global explanations of the proposed method and TFT, it is evident that TFT exhibited a distribution of contributions that lacks clarity across various variables and time points, making it challenging to interpret the rationale behind the predictions.
By contrast, the proposed method highlighted the contribution of a smaller number of patches.
Specifically, the patch immediately preceding the prediction of the target electricity price (y) as well as the future patches for system load (exogenous 1) and generation (exogenous 2), strongly contributed to the predictions.

\subsubsection{Quantitative Evaluation.}
\begin{figure}[htbp]
    \centering
    \includegraphics[keepaspectratio, width=\linewidth]{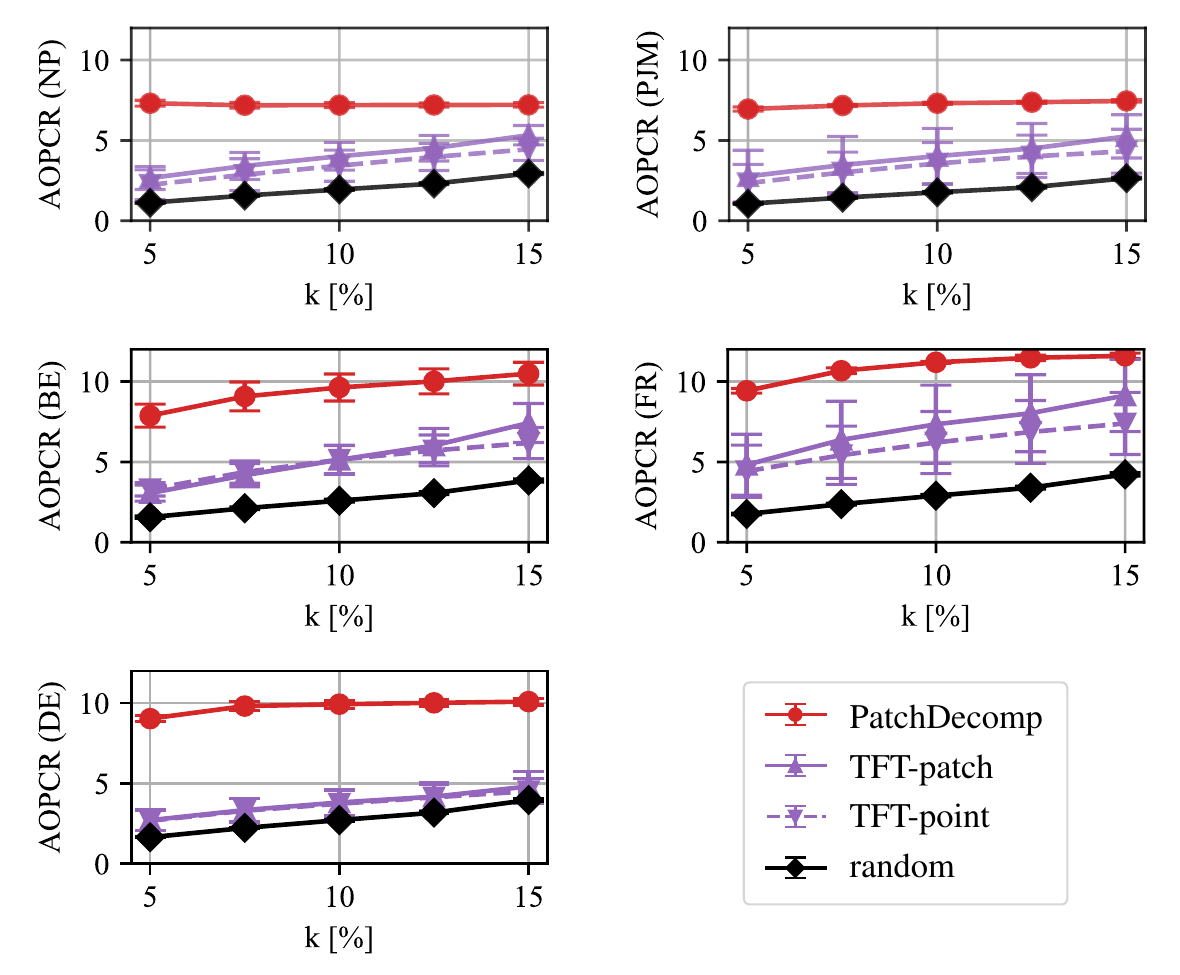}
    \caption{$AOPCR_k$ for $k \in \{5.0, 7.5, 10.0, 12.5, 15.0\}$. The error bars represent the standard deviation.}
    \label{fig:AOPCR}
\end{figure}

To quantitatively evaluate the interpretability of TSF models, we introduce the concept of comprehensiveness~\cite{deyoung2019eraser}.
Comprehensiveness measures how much the output changes when the most important $k\%$ of features from the input are removed; the greater the informational value of the removed input for the output, the larger this metric becomes.
In practice, we quantified comprehensiveness using the area over the perturbation curve for regression (AOPCR)~\cite{ozyegen2022evaluation}.
By applying MAE as a benchmark, it can be expressed as follows:
\begin{equation}
    AOPCR_{k}=\frac{\sum_t^T\text{MAE}\left(\mathcal{F}(\bm{x}_t), \mathcal{F}(\bm{x}_{t,\backslash k})\right)}{TH}.
\end{equation}
Here, $\mathcal{F}$ represents the forecasting model, $\bm{x}_t$ denotes the $t$-th input, and $\bm{x}_{t,\backslash k}$ refers to the input after removing the top $k\%$ of values with high prediction contributions from the $t$-th input.
$T$ is the length of the test data and $H$ is the prediction horizon.
In this experiment, we set $K = \{5.0, 7.5, 10.0, 12.5, 15.0\}$ and $k \in K$.
For each $k$, we computed the values individually, and we did not calculate the average $AOPCR = \frac{1}{|K|}\sum_k AOPCR_k$ across the entire set $K$.
Additionally, we removed input values by replacing the values in the patches with the mean values of the variables in the entire test data.
We calculated $AOPCR_k$ in four different ways: removing patches with high contributions from PatchDecomp (PatchDecomp), removing random patches from PatchDecomp (random), removing point-wise inputs from TFT (TFT-point), and removing inputs aggregated by patch-wise importance from TFT (TFT-patch) (Figure \ref{fig:AOPCR}).
In all five market datasets, PatchDecomp outperformed both the random and TFT values.
This result quantitatively demonstrates that the proposed method can effectively highlight patches with a high contribution to the predictions.

\section{Conclusion}
In this paper, we introduced PatchDecomp, a TSF method that decomposes predicted values into contributions of the input subsequences, explaining how much each subsequence of any variable, including exogenous variables, contributes to the prediction.
We conducted experiments on an LTSF task, which consisted of seven types of data, and an electricity price forecasting task, which included five market datasets with exogenous variables.
The results demonstrated that the proposed method can achieve a prediction accuracy comparable to that of recently proposed forecasting methods.
Furthermore, through the visualization of prediction contributions and quantitative evaluation using AOPCR, we clarified that the interpretability of PatchDecomp was improved compared with that of TFT, which is also an interpretable model.

Currently, PatchDecomp cannot present the contributions of static exogenous variables to predictions, and we plan to address this issue in future studies.

\section*{Acknowledgments}
We are grateful to Professor Yasushi Sakurai and Professor Yasuko Matsubara  of the University of Osaka for their helpful discussions and valuable advice regarding this study.

\bibliography{citation}

\section{Appendix}

\subsection{LTSF Datasets Description}
A description of the long-term time series forecasting datasets is presented in Table \ref{tab:ltsf_datasets}.

\subsubsection{Electricity Transformer Temperature (ETT).}
The data from two locations in China, collected from July 2016 to July 2018, are referred to as ETTh1 and ETTh2 for data sampled at hourly intervals, and ETTm1 and ETTm2 for data sampled at 15 min intervals~\cite{zhou2021informer}.
\subsubsection{Weather.}
This dataset consists of meteorological measurements from the Weather Station of the Max Planck Biogeochemistry Institute in Germany for the year 2020~\cite{wu2021autoformer}.
\subsubsection{Electricity Consuming Load (ECL).}
The dataset consists of power consumption (kWh) collected every 15 min for each client from 2012 to 2014, which has been aggregated into hourly data~\cite{li2019enhancing}.
\subsubsection{Traffic.}
The dataset contains road occupancy rates (ranging from 0 to 1) for general roads in the San Francisco Bay Area, collected from January 2015 to December 2016~\cite{lai2018modeling,wu2021autoformer}.

\begin{table*}[htbp]
    \centering
    \small
    \begin{tabular}{c|c|c|c}
        \toprule
         Dataset & Dim & Sampling Frequency & (Train, Valid, Test) \\
        \midrule
        ETTh1 & 7 & 1 h & (8640, 2880, 2880) \\
        \midrule
        ETTh2 & 7 & 1 h & (8640, 2880, 2880) \\
        \midrule
        ETTm1 & 7 & 15 min & (34560, 11520, 11520) \\
        \midrule
        ETTm2 & 7 & 15 min & (34560, 11520, 11520) \\
        \midrule
        Weather & 21 & 10 min & (36887, 10539, 5269) \\
        \midrule
        ECL & 321 & 1 h & (18414, 5260, 2630) \\
        \midrule
        Traffic & 862 & 1 h & (12282, 3508, 1754) \\
        \bottomrule
    \end{tabular}
    \caption{Description of LTSF datasets}
    \label{tab:ltsf_datasets}
\end{table*}

\subsection{EPF Datasets Description}
This dataset consists of actual electricity prices from five electricity markets (NP, PJM, BE, FR, and DE)~\cite{lago2021forecasting}.
The description of the datasets is summarized in Table \ref{tab:epf_datasets}.

\begin{table*}[htbp]
    \centering
    \small
    \begin{tabular}{c|p{4cm}|p{3cm}|p{3cm}|p{2cm}|l}
        \toprule
        & Market & Exogenous variable 1 & Exogenous variable 2 & Period & (Train, Valid, Test) \\
        \midrule
        NP & the Pennsylvania-New Jersey-Maryland market & 2 day-ahead system load & day-ahead wind generation & 01-01-2013 to 24-12-2018 & (36504, 5448, 10464) \\
        \midrule
        PJM & the Nord Pool market & day-ahead load & 2 day-ahead COMED load & 01-01-2013 to 24-12-2018 & (36504, 5448, 10464) \\
        \midrule
        BE & the Belgium markets & day-ahead load & day-ahead total France generation & 09-01-2011 to 31-12-2016 & (36504, 5448, 10464) \\
        \midrule
        FR & the France markets & day-ahead load & day-ahead total France generation & 09-01-2011 to 31-12-2016 & (36504, 5448, 10464) \\
        \midrule
        DE & the Germany markets & day-ahead zonal load & day-ahead wind and solar generation & 09-01-2012 to 31-12-2017 & (36504, 5448, 10464) \\
        \bottomrule
    \end{tabular}
    \caption{Description of EPF datasets}
    \label{tab:epf_datasets}
\end{table*}

\subsection{Hyperparameters for LSTF}
The search range for the hyperparameters is summarized in Table \ref{tab:ltsf_hyperparameters}.
The final selected hyperparameters varied depending on the trials.

\begin{table*}[htbp]
    \centering
    \small
    \begin{tabular}{c|c|c|c|c|c|c|c|c}
        \toprule
        & Patch Size & Hidden Size & Heads & Layers & Units & Blocks & Window Size & Dropout \\
        \midrule
        PatchDecomp & 12,24,48 & 32,64,128,256 & 4,8 & 1,2,3,4 & - & - & - & uniform(0.0,0.5) \\
        \midrule
        PatchTST & 12,24,48 & 32,64,128,256 & 4,8 & - & - & - & - & uniform(0.0,0.5) \\
        \midrule
        NBEATSx & - & - & - & - & 32,64,128,256 & - & - & uniform(0.0,0.5) \\
        \midrule
        NHITS & - & - & - & - & 32,64,128,256 & - & - & uniform(0.0,0.5) \\
        \midrule
        TFT & - & 32,64,128,256 & 4,8 & - & - & - & - & uniform(0.0,0.5) \\
        \midrule
        DLinear & - & - & - & - & - & - & 11,25,51 & - \\
        \midrule
        TSMixer & - & 32,64,128,256 & - & - & - & 1,2,4,6,8 & - & uniform(0.0,0.5) \\
        \midrule
        Autoformer & - & 32,64,128,256 & 4,8 & - & - & - & - & uniform(0.0,0.5) \\
        \midrule
        iTransformer & - & 32,64,128,256 & 4,8 & 1,2,3,4 & - & - & - & uniform(0.0,0.5) \\
        \midrule
        TiDE & - & 32,64,128,256 & - & 1,2,3,4 & - & - & - & uniform(0.0, 0.5) \\
        \bottomrule
    \end{tabular}
    \caption{Hyperparameters tuning spaces for LTSF}
    \label{tab:ltsf_hyperparameters}
\end{table*}

\subsection{Hyperparameters for EPF}
The search range for the hyperparameters is summarized in Table \ref{tab:epf_hyperparameters}.
The final selected hyperparameters varied depending on the trials.

\begin{table*}[htbp]
    \centering
    \small
    \begin{tabular}{c|c|c|c|c|c|c}
        \toprule
        & Hidden Size & Heads & Layers & Units & Blocks & Dropout \\
        \midrule
        PatchDecomp & 16,32,64,128,256,512 & 4,8 & 1,2,3,4 & - & - & uniform(0.0,0.5) \\
        \midrule
        NBEATSx & - & - & - & 16,32,64,128,256,512 & - & uniform(0.0,0.5) \\
        \midrule
        NHITS & - & - & - & 16,32,64,128,256,512 & - & uniform(0.0,0.5) \\
        \midrule
        TFT & 16,32,64,128,256,512 & 4,8 & - & - & - & uniform(0.0,0.5) \\
        \midrule
        TSMixer & 16,32,64,128,256,512 & - & - & - & 1,2,4,6,8 & uniform(0.0,0.5) \\
        \midrule
        TiDE & 16,32,64,128,256,512 & - & 1,2,3,4 & - & - & uniform(0.0, 0.5) \\
        \bottomrule
    \end{tabular}
    \caption{Hyperparameters tuning spaces for EPF}
    \label{tab:epf_hyperparameters}
\end{table*}

\subsection{Experimental Details}
In LTSF, the batch size was set to 8, while the windows batch size was set to 128.
In EPF, these values were set to 16 and 512, respectively.
However, owing to memory constraints, the windows batch size was set to 64 specifically for TFT, which requires significantly more memory, in LTSF.
For the same reason, TSMixer was also configured with a batch size of 4 instead of 8 in some LTSF experiments.
The random seed values were specified within the program at the start of each experiment.
Hyperparameter optimization using Optuna~\cite{akiba2019optuna} was conducted with a maximum of 20 epochs for LTSF, terminating the training if no improvement in the evaluation metric was observed over 5 epochs.
For EPF, the maximum limit was set to 2000 epochs, and training was stopped if no improvement was seen over 20 epochs.

\subsection{Global Explanations}
\begin{figure*}[htbp]
    \centering
    \includegraphics[keepaspectratio, width=\linewidth]{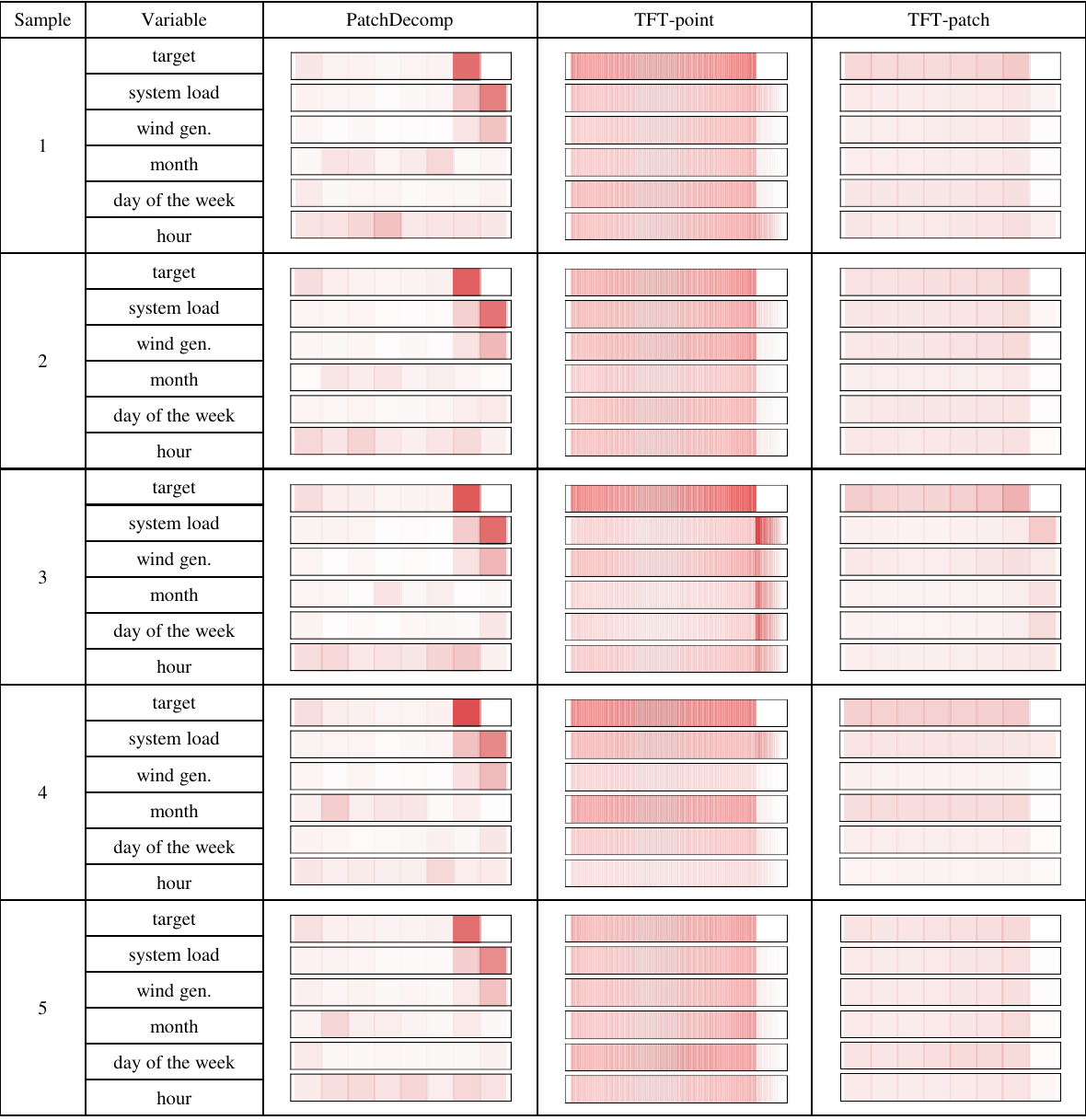}
    \caption{Global explanations on the NP dataset. The term ``gen." indicates ``generation."}
    \label{fig:global_explanations_NP}
\end{figure*}

\begin{figure*}[htbp]
    \centering
    \includegraphics[keepaspectratio, width=\linewidth]{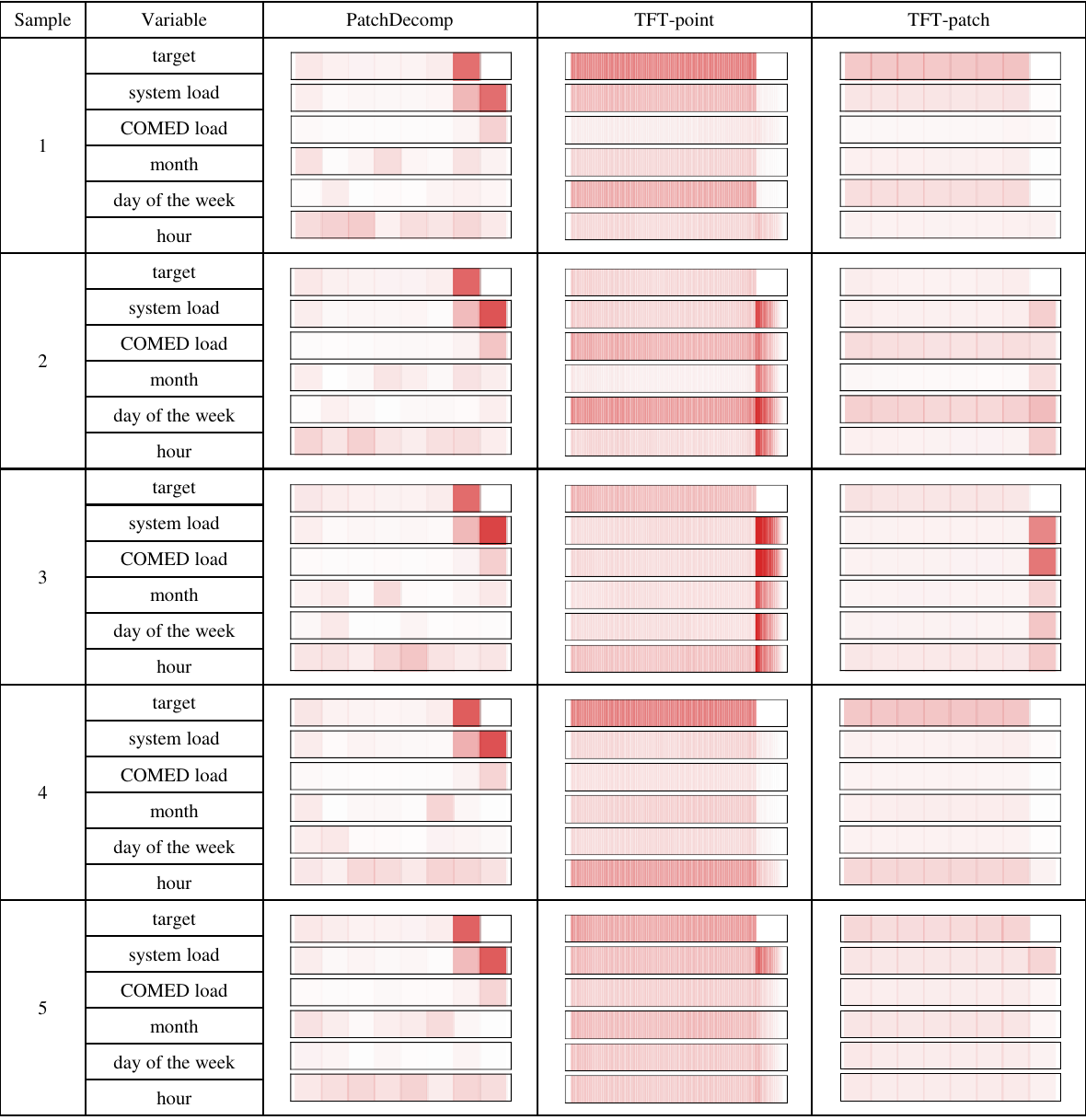}
    \caption{Global explanations on the PJM dataset}
    \label{fig:global_explanations_PJM}
\end{figure*}

\begin{figure*}[htbp]
    \centering
    \includegraphics[keepaspectratio, width=\linewidth]{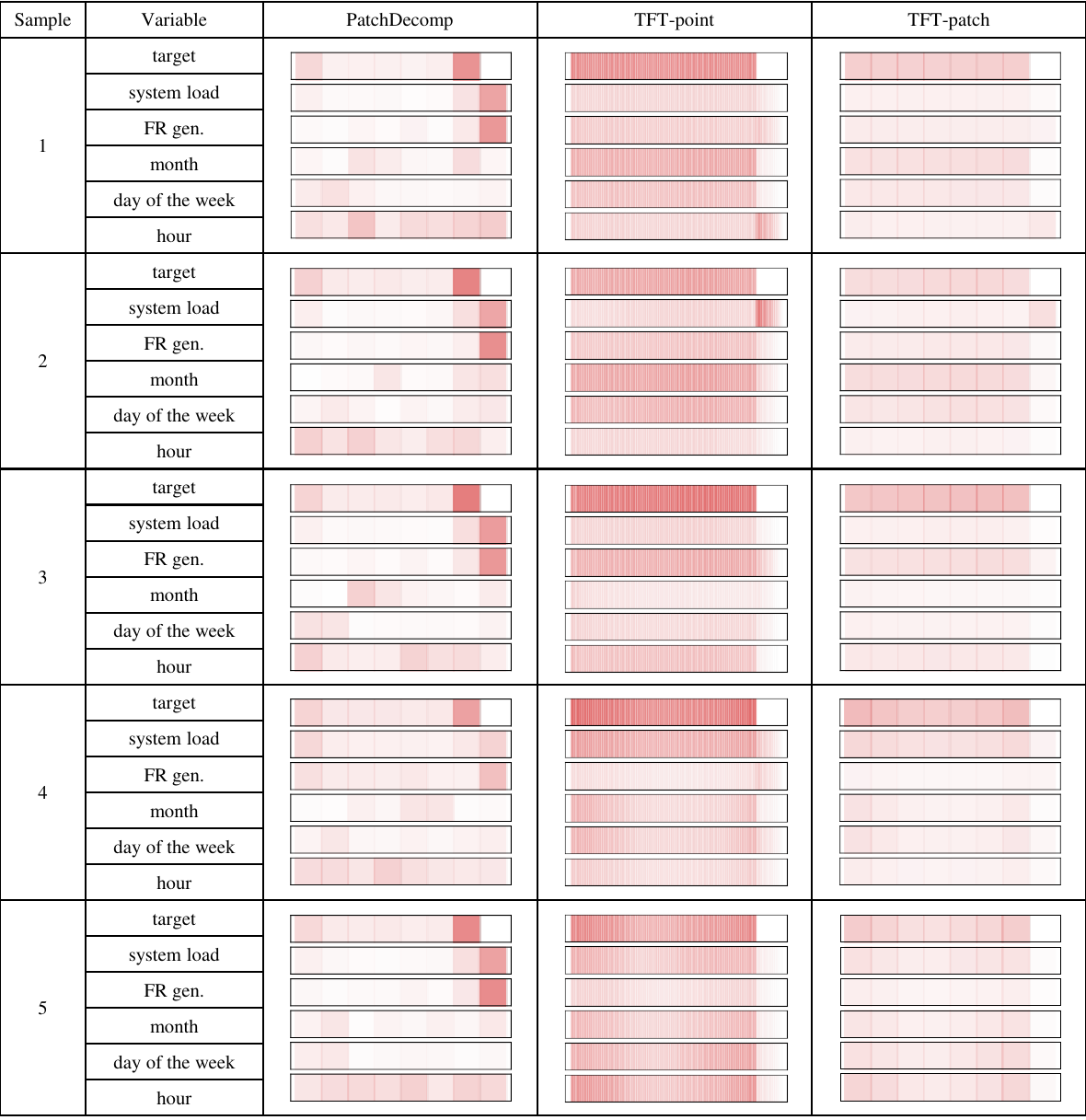}
    \caption{Global explanations on the BE dataset. The term ``gen." indicates ``generation."}
    \label{fig:global_explanations_BE}
\end{figure*}

\begin{figure*}[htbp]
    \centering
    \includegraphics[keepaspectratio, width=\linewidth]{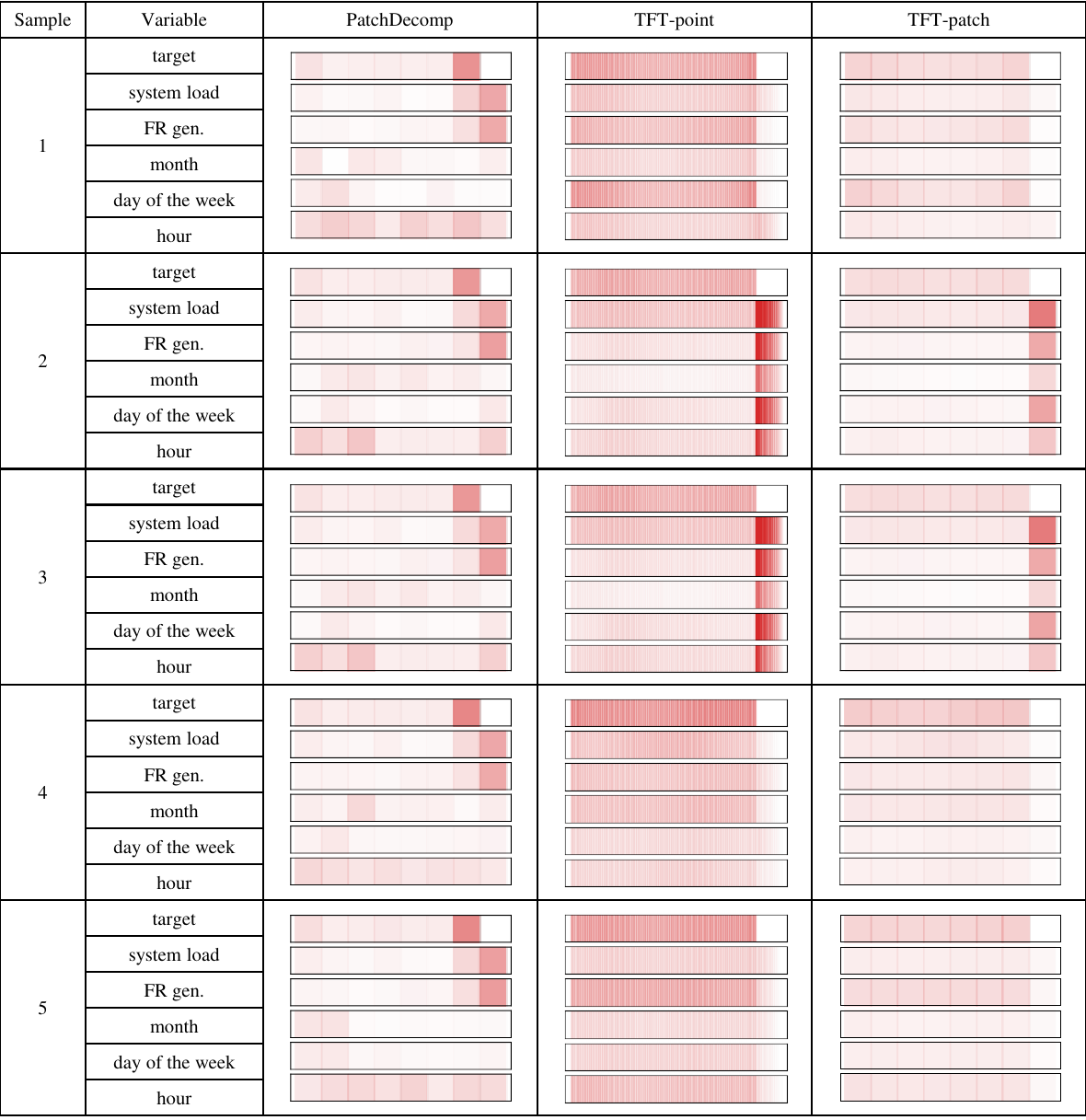}
    \caption{Global explanations on the FR dataset. The term ``gen." indicates ``generation."}
    \label{fig:global_explanations_FR}
\end{figure*}

\begin{figure*}[htbp]
    \centering
    \includegraphics[keepaspectratio, width=\linewidth]{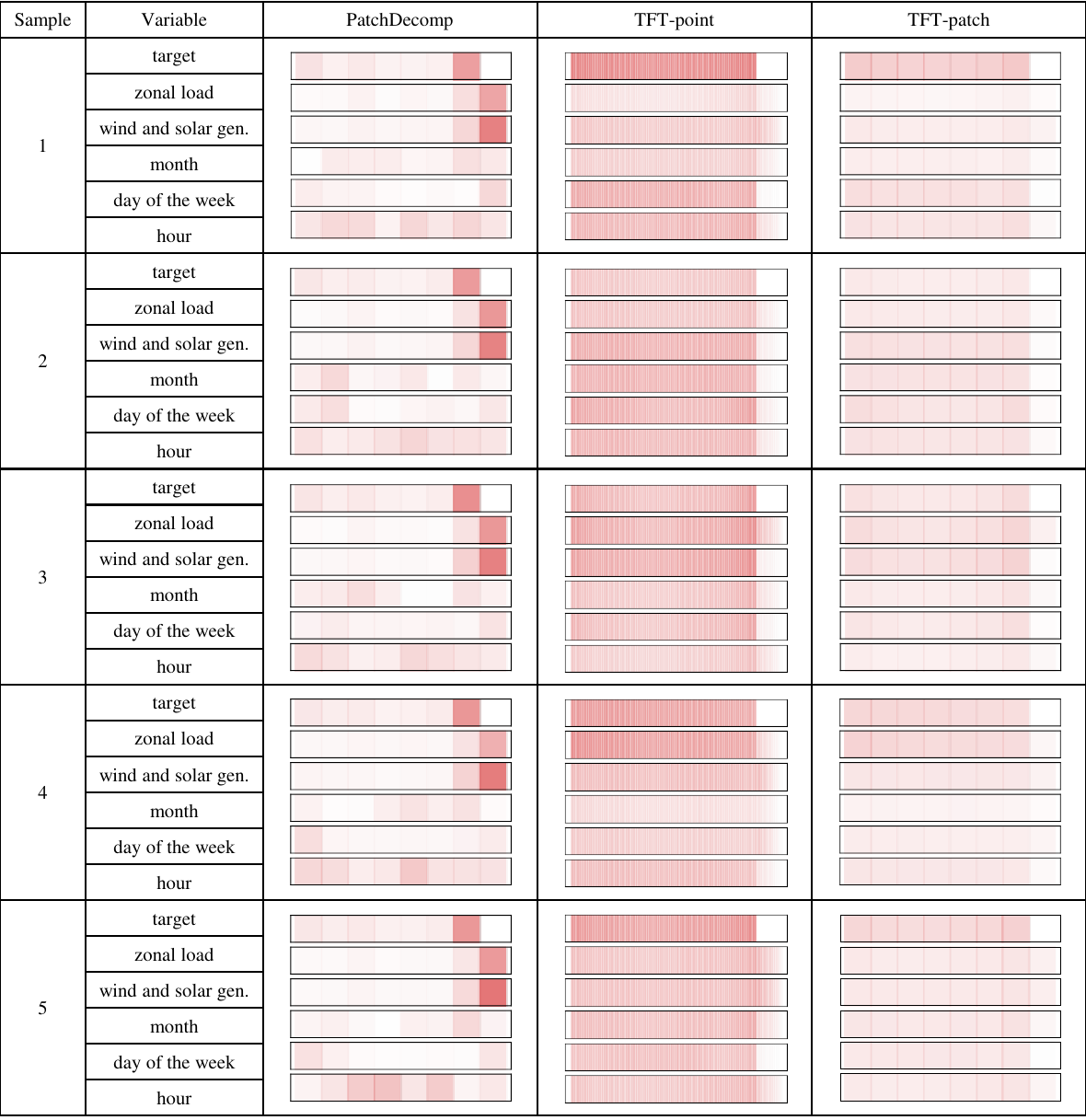}
    \caption{Global explanations on the DE dataset. The term ``gen." indicates ``generation."}
    \label{fig:global_explanations_DE}
\end{figure*}

We conducted five trials using different random seeds for each of the five market datasets in the EPF dataset.
Figure \ref{fig:global_explanations_NP}–\ref{fig:global_explanations_DE} show the global explanations from the five trials of PatchDecomp and TFT.
The TFT-point visualizes the predicted contributions calculated at each time step, whereas TFT-patch aggregates and visualizes the contributions at the patch level.
PatchDecomp emphasized approximately the same patches when trained on the same market data, even when the models were initialized with different random seeds.
By contrast, TFT exhibited a less distinct distribution of predicted contributions, with significant variations in the contributions of certain market data across trials.
These results indicate that PatchDecomp provides a higher and more robust level of interpretability than TFT.

\end{document}